\documentclass[conference]{IEEEtran}
\IEEEoverridecommandlockouts
\usepackage{cite}
\usepackage{amsmath,amssymb,amsfonts}
\usepackage{algorithmic}
\usepackage{graphicx}
\usepackage{textcomp}
\usepackage{xcolor}
\usepackage{multirow}
\usepackage{cellspace}
\usepackage{makecell}
\usepackage{url}
\def\BibTeX{{\rm B\kern-.05em{\sc i\kern-.025em b}\kern-.08em
    T\kern-.1667em\lower.7ex\hbox{E}\kern-.125emX}}

\begin{document}
\title{DSAT-HD: Dual-Stream Adaptive Transformer with Hybrid Decomposition for Multivariate Time Series Forecasting\\}

\author{\IEEEauthorblockN{1\textsuperscript{st} Zixu Wang}
\IEEEauthorblockA{\textit{Computer Science and Technology} \\
\textit{Harbin Engineering University}\\
Harbin, China \\
wangzixu@hrbeu.edu.cn}
\and
\IEEEauthorblockN{2\textsuperscript{nd} Hongbin Dong\textsuperscript{*}}
\IEEEauthorblockA{\textit{Computer Science and Technology} \\
\textit{Harbin Engineering University}\\
Harbin, China \\
donghongbin@hrbeu.edu.cn}
\and
\IEEEauthorblockN{3\textsuperscript{rd} Xiaoping Zhang}
\IEEEauthorblockA{\textit{Traditional Chinese Medicine Data Center} \\
\textit{China Academy of Chinese Medical Sciences}\\
Beijing, China \\
zhangxp@ndctcm.cn}
}
\maketitle

\renewcommand{\thefootnote}{}{
    \footnotetext{
        This work is supported by the National Natural Science Foundation of China under Grant No. 82374621.
    }
}

\begin{abstract}
Time series forecasting is crucial for various applications, such as weather, traffic, electricity, and energy predictions. Currently, common time series forecasting methods are based on Transformers. However, existing approaches primarily model limited time series or fixed scales, making it more challenging to capture diverse features across different ranges. Additionally, traditional methods like STL for complex seasonality-trend decomposition require pre-specified seasonal periods and typically handle only single, fixed seasonality. We propose the Hybrid Decomposition Dual-Stream Adaptive Transformer (DSAT-HD), which integrates three key innovations to address the limitations of existing methods: 1) A hybrid decomposition mechanism combining EMA and Fourier decomposition with RevIN normalization, dynamically balancing seasonal and trend components through noise Top-k gating; 2) A multi-scale adaptive pathway leveraging a sparse allocator to route features to four parallel Transformer layers, followed by feature merging via a sparse combiner, enhanced by hybrid attention combining local CNNs and global interactions; 3) A dual-stream residual learning framework where CNN and MLP branches separately process seasonal and trend components, coordinated by a balanced loss function minimizing expert collaboration variance. Extensive experiments on nine datasets demonstrate that DSAT-HD outperforms existing methods overall and achieves state-of-the-art performance on some datasets. Notably, it also exhibits stronger generalization capabilities across various transfer scenarios.
\end{abstract}


\begin{figure}[htbp]
    \centering
    \includegraphics[width=0.75\linewidth]{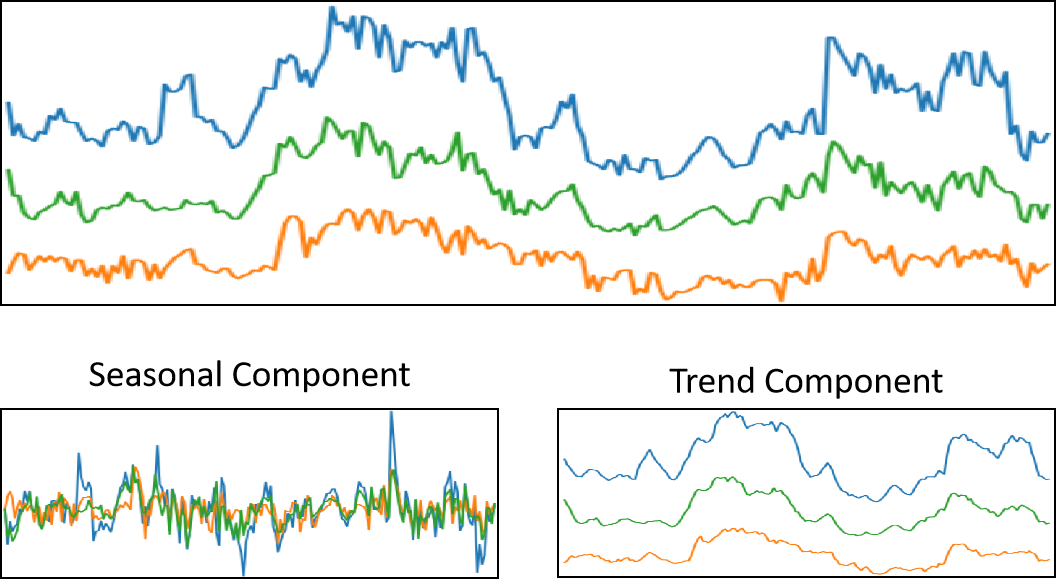}
    \caption{Season-Trend Decomposition of the Original Time Series}
    \label{fig:seasonTrendDecomp}
\end{figure}

\section{Introduction}
Time series forecasting holds significant application value in various fields such as energy management and financial analysis. Inspired by the success of Transformer models in domains like computer vision and natural language processing, Transformer-based methods for time series forecasting have recently garnered widespread attention. However, effectively capturing the season-trend interaction patterns and multi-scale characteristics within time series remains a significant challenge. Traditional decomposition methods like STL \cite{1990STL} exhibit limitations when handling non-stationary sequences, while many existing Transformer variants \cite{vaswani2017attention, kitaev2020reformer, cai2024mambats} still have room for improvement in the accuracy of long-term forecasting tasks. Therefore, further exploring the potential of Transformers in time series forecasting is of great importance.

Although Transformer models have significantly improved the performance of long-sequence forecasting tasks \cite{lim2021temporal,ma2024fmamba} and demonstrated a powerful capacity for capturing long-range dependencies, directly applying them to time series data presents several challenges. Real-world time series often possess multi-scale properties (e.g., daily, weekly, yearly cycles) and typically exhibit non-stationarity. While some studies have attempted to mitigate distribution shift issues using techniques like Reversible Instance Normalization (RevIN) \cite{guo2024weits}, effectively separating and modeling the dynamic interactions between trend and seasonal components within an end-to-end learning framework remains an open problem. For instance, the recently proposed Non-stationary Transformers \cite{liu2022non} reduced MSE by 49\%in electricity load forecasting tasks through a de-stationary attention mechanism. Similarly, dynamic frequency domain decomposition frameworks \cite{ye2024frequency} have achieved breakthroughs in modeling non-stationary sequences.

\begin{figure*}
\centering
\includegraphics[width=1\linewidth]{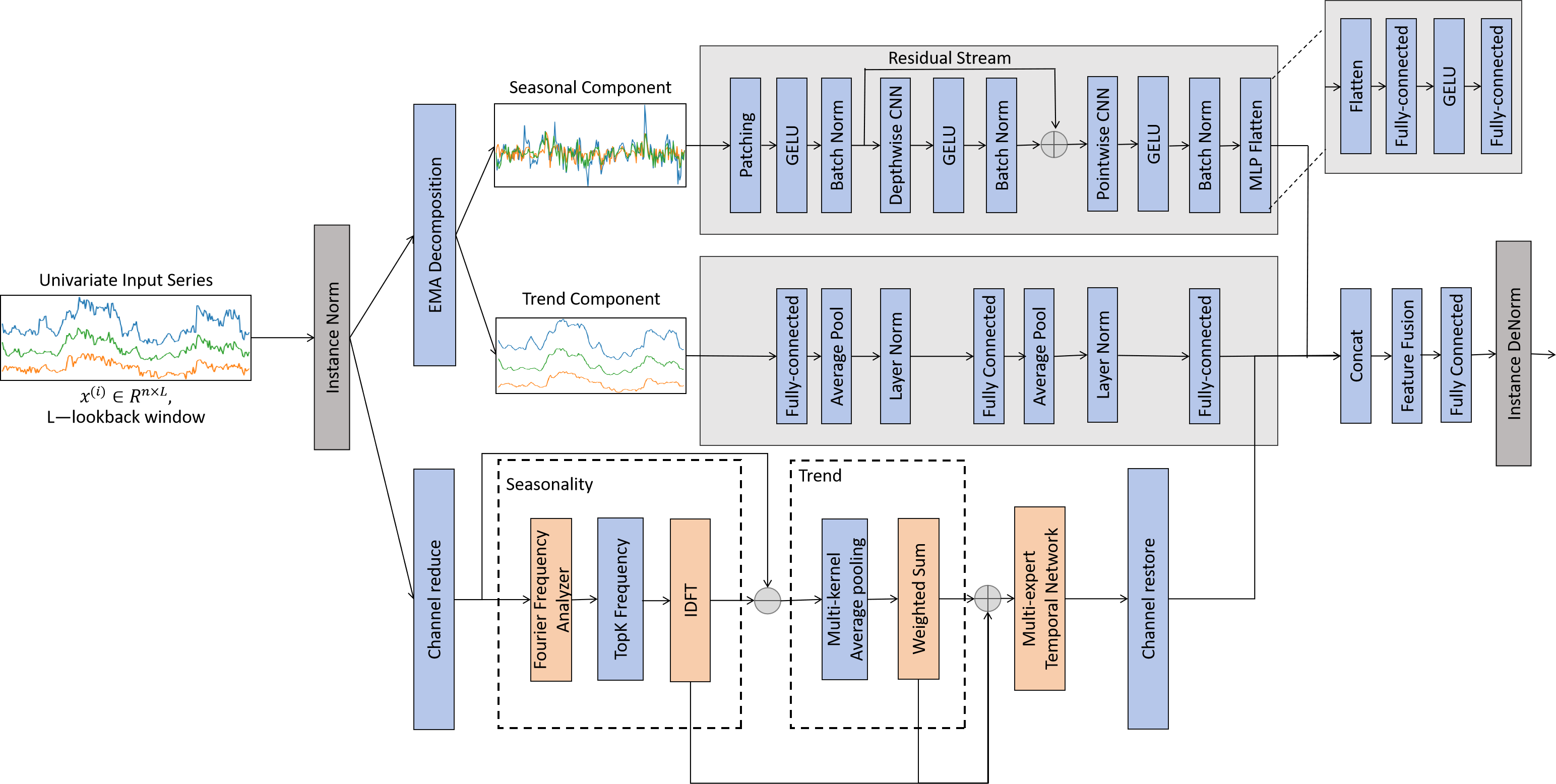}
\caption{Overview of the DSAT-HD Model: After normalization, the input time series is processed through two parallel paths. One path undergoes EMA decomposition in the time domain, separating the series into seasonal and trend components, which are then processed by a dual-stream network. The other path performs frequency-domain feature extraction on the sequence.}
\label{fig:model_overall}
\end{figure*}

On the other hand, the multi-scale characteristics of time series require models to have the ability to extract information across different resolutions. Many existing methods rely on fixed segment partitioning strategies \cite{2021Autoformer,kumar2024spatio} or a single attention mechanism \cite{zhang2023crossformer, hanshuang21}, which limits their ability to simultaneously capture short-term fluctuations and long-term evolutions. To address this challenge, Pathformer \cite{chen2024pathformer} proposed an adaptive path mechanism, which, through multi-scale partitioning and dual-attention fusion, reduced periodic forecasting error by 23\% across 11 benchmark datasets. Furthermore, the computational complexity of the standard self-attention mechanism grows quadratically with sequence length, posing an efficiency bottleneck for processing long sequences. VARMAformer \cite{song2025varma} enhanced inference speed by 4.8 times while maintaining state-of-the-art (SOTA) accuracy on the ETTh1 dataset by integrating classical statistical methods with modern architectural design.

Additionally, the Hierarchical Transformer \cite{liu2024hidformer} has demonstrated advantages in long-horizon forecasting tasks through hierarchical temporal modeling. Mixed Frequency Transformers \cite{wang2025mixed} innovatively addressed the challenge of fusing multi-frequency data in economic forecasting. For complex spatio-temporal interaction scenarios, Spatio-temporal Decomposition \cite{jin2023spatio} combined graph neural networks with temporal decomposition to achieve breakthroughs in traffic flow prediction accuracy.

To address the aforementioned challenges, we propose a novel framework named DSAT-HD (Dual-Stream Adaptive Transformer with Hybrid Decomposition). Our work primarily makes the following contributions:

\begin{figure*}
\centering
\includegraphics[width=1\linewidth]{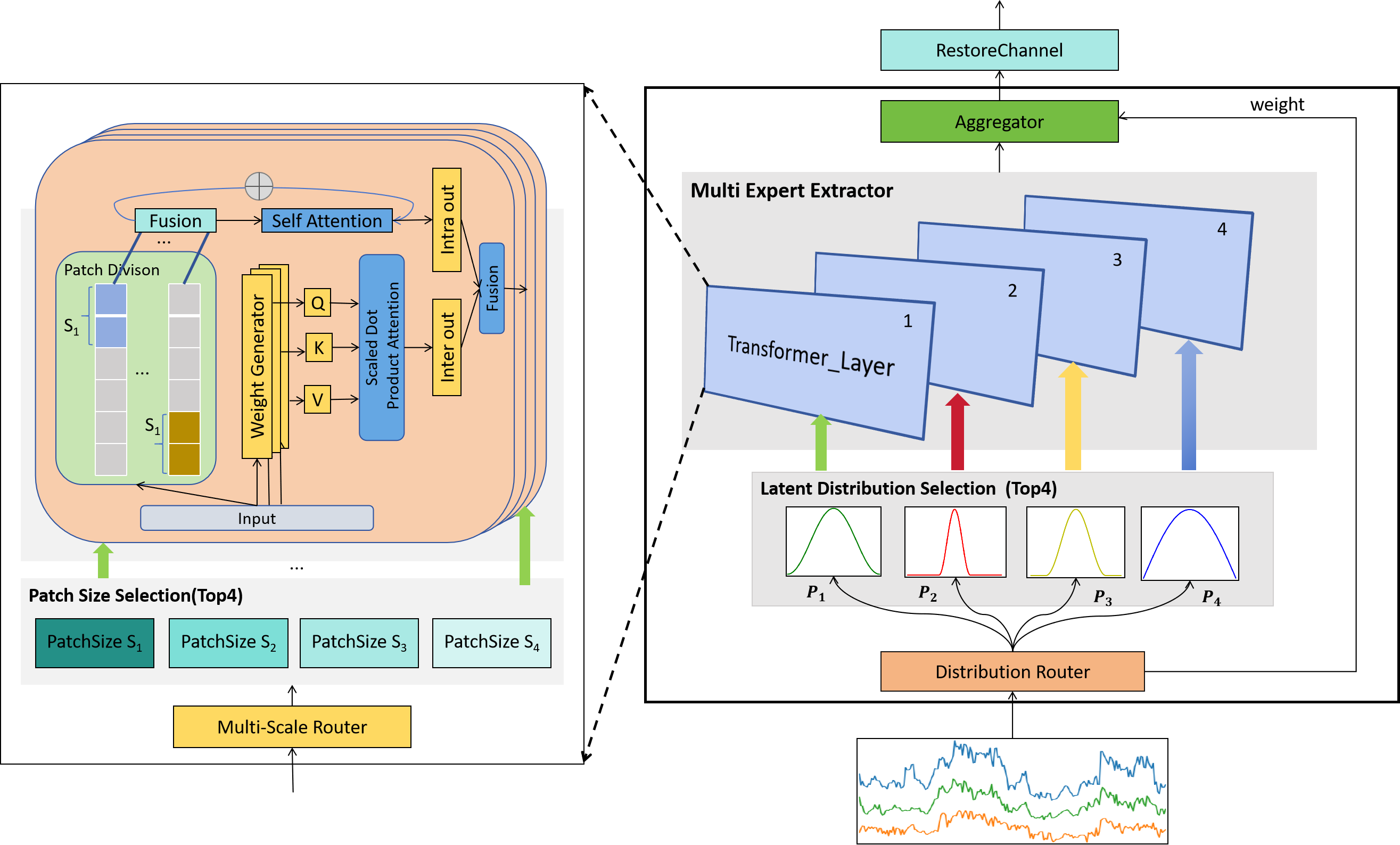}
\caption{The adaptive routing mechanism selects the top 4 weight values through a sparse dispatcher to dynamically choose multi-scale feature patches. Within the four-layer multi-expert architecture, each layer specializes in processing patches of specific sizes, achieving cross-scale feature interaction through the fusion of global attention and inter-patch local attention.}
\label{fig:Transformer_layer}
\end{figure*}

\begin{enumerate}

\item We introduce a hybrid adaptive decomposition module. It integrates EMA decomposition, Fourier decomposition, and a multi-scale moving average module, dynamically balancing seasonal and trend components via a noisy Top-k gating mechanism.

\item We design a multi-scale adaptive path architecture. It employs a Sparse Dispatcher to dynamically route features to 4 Transformer layers with different patch sizes, combining the local attention of Convolutional Neural Networks (CNN) with the global interaction capability of the Transformer (Fig.~\ref{fig:Transformer_layer}).

\item We construct a dual-stream residual learning framework. It consists of two parts: a CNN stream for processing seasonal components and an MLP stream for capturing trend components (Fig.~\ref{fig:model_overall}). We also introduce a carefully designed Balance Loss function to minimize the collaboration variance among experts.

\end{enumerate}

\section{Related work}

\subsection{Time series decomposition method}

Time series decomposition is a crucial preprocessing step in forecasting tasks. Traditional methods like STL (Season-Trend decomposition using Loess) \cite{1990STL} decompose a time series into three primary components: trend, seasonal, and residual. STL uses LOESS (locally estimated scatterplot smoothing) to extract smooth estimates of these three components. However, it performs poorly when handling time series with multiple seasonal patterns or non-fixed periods. In recent years, deep learning modules have begun to integrate decomposition in an end-to-end manner. For instance, N-BEATS \cite{oreshkin2019n} employs a dual residual stacking structure to learn the seasonal and trend parts separately; DeepAR \cite{salinas2020deepar} implicitly learns the decomposition through an autoregressive model; Autoformer \cite{2021Autoformer} introduces a series decomposition block to achieve adaptive separation.

Although these methods are efficient, they are limited by relying on a single decomposition mechanism. Our work innovatively combines EMA smoothing \cite{1985Exponential}, Fourier decomposition, and multi-scale moving averages, achieving dynamic balancing of the model components via a learnable gating mechanism.

\subsection{Transformer-based Forecasting Models}

Transformer models often demonstrate significant advantages in handling long-sequence forecasting tasks. For instance, Informer \cite{zhou2021informer} reduces computational complexity through ProbSparse attention; FEDformer \cite{zhou2022fedformer} incorporates Fourier and wavelet transforms to enhance frequency-domain modeling; PatchTST \cite{nie2022time} employs a patching strategy to improve cross-channel generalization capability. However, these methods exhibit limitations when processing multi-scale features:

\begin{enumerate}
    \item Fixed patch sizes struggle to adapt to different temporal granularities (e.g., daily vs. seasonal scales).
    \item Global attention mechanisms can easily overlook certain local features.
\end{enumerate}

Different from the aforementioned approaches, we propose a multi-scale adaptive path architecture. This architecture utilizes a Sparse Dispatcher to dynamically route features to Transformer layers of varying scales, and incorporates CNNs to enhance the extraction of fine-grained features.

\subsection{Multi-scale Modeling and Feature Fusion}

Multi-scale modeling can effectively capture the hierarchical features of time series. MTGNN \cite{wu2020connecting} models spatio-temporal dependencies through graph neural networks; Scaleformer \cite{shabani2022scaleformer} employs a hierarchical pyramid structure to handle different temporal resolutions; StemGNN \cite{cao2020spectral} combines spectral graph convolution to capture frequency-domain features. Regarding feature fusion, the Mixture of Experts (MoE) model integrates expert outputs via a gating mechanism, but it faces the challenge of expert load imbalance.

\section{Preliminaries}

The input for multivariate time series forecasting is a historical sequence of length $L$, denoted as $x=(x_1,x_2,\dots,x_L)$. The task is to predict the future $T$-step sequence $\hat{x}=(x_{L+1},x_{L+2},\dots,x_{L+T})$, where $x_t$ represents an $M$-dimensional vector at time $t$. Consequently, the input of the multivariate time series can be represented as $x\in \mathbb{R}^{M\times L}$, andd the output can be represented as $\hat{x}\in \mathbb{R}^{M\times T}$.

\subsection{Seasonal-Trend Decomposition}

Seasonal-trend decomposition is commonly employed as a preprocessing step for model inputs. Decomposing a time series into seasonal and trend components enhances the model's ability to learn complex temporal patterns. Fig.~\ref{fig:seasonTrendDecomp} illustrates the decomposition of a sample series from the ETTh1 dataset into its seasonal and trend series. The trend component captures the long-term evolution direction of the data, which typically exhibits linear characteristics. In contrast, the seasonal component is designed to capture patterns or cycles that occur at regular intervals; these variations are often periodic and nonlinear. By separately learning these two distinct components and subsequently integrating their features, the model can more effectively generate accurate forecasts.

\textbf{Exponential Moving Average} \cite{1985Exponential} is a weighted average method that assigns greater importance to more recent data points. Given a sequence of nweighted data points $[\theta_1,\theta_2,\dots,\theta_n]$, the EMA is calculated as follows:

\begin{align}
v_0 &= 0 \\
v_t &= \alpha v_{t-1} + (1-\alpha)\theta_t, \quad t>0 \\
X_T &= \text{EMA}(X) \\
X_S &= X - X_T
\end{align}

where $\alpha$ is the smoothing factor, with $0<\alpha<1$.

\section{Methodology}

\begin{figure*}[h]
\centering
\includegraphics[width=1\textwidth, keepaspectratio]{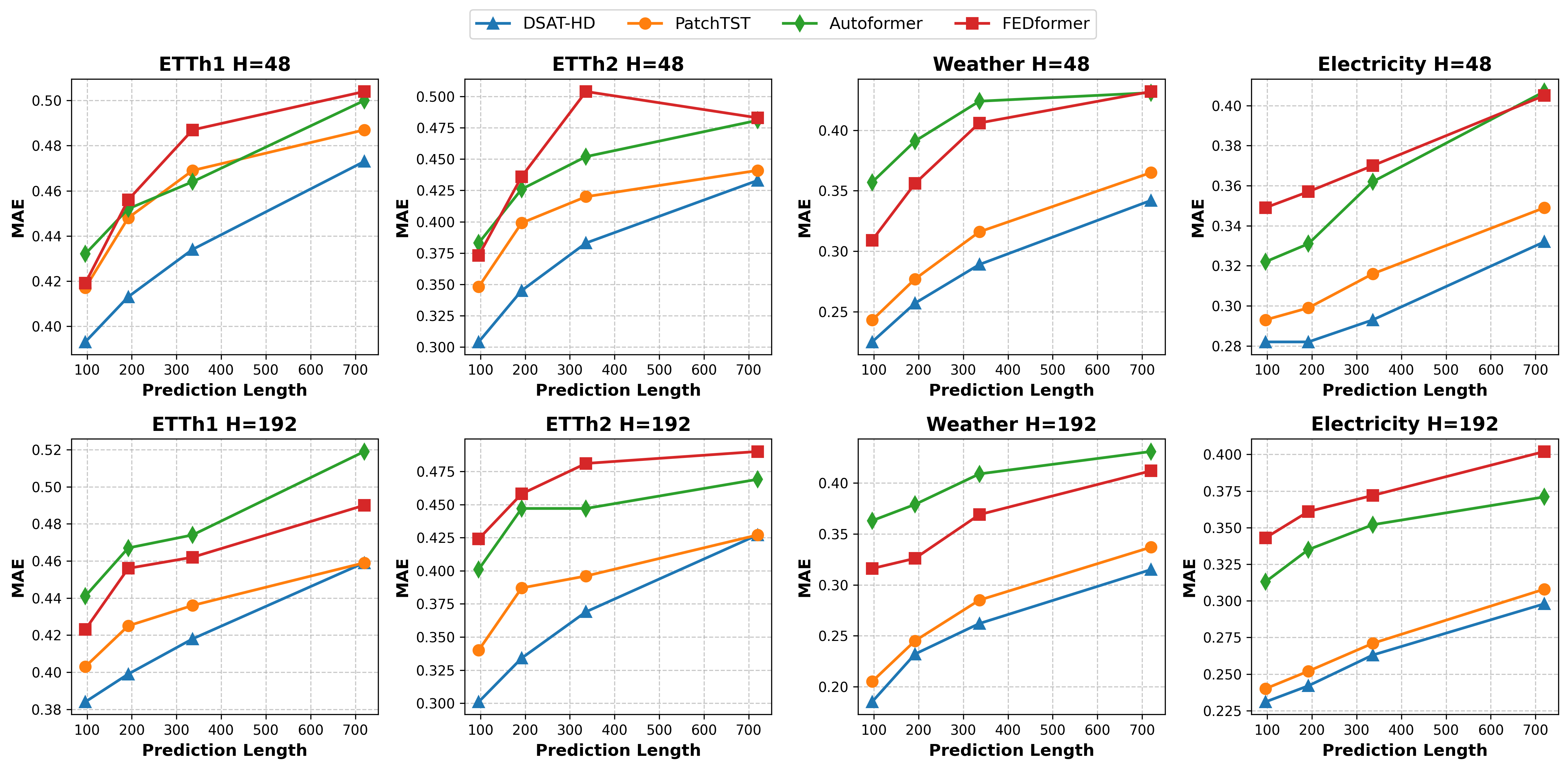}
\caption{Comparative Analysis of Forecasting Results under Different Input Lengths on ETTh1, ETTh2, Weather, and Electricity Datasets}
\label{fig:48和192长度比较}
\end{figure*}

\subsection{Structural Overview}

Fig.~\ref{fig:model_overall} illustrates the architecture of the DSAT-HD model, which explores the latent features of sequences simultaneously in both the frequency and time domains. Specifically, we first apply Instance Normalization to align the distributions of the training and test data. Subsequently, the EMA module decomposes the data into seasonal (S) and trend (T) components, which are processed separately by a dual-stream network. To prevent excessively high data dimensionality, another branch of the data undergoes dimensionality reduction via a fully connected layer. This is followed by a Fourier transform to convert the time-domain data into the frequency domain. Seasonal series are obtained via Top-K selection, and the trend series is derived through multi-kernel average pooling. A specially designed distribution routing mechanism then directs time series with similar latent distributions to expert networks based on multiple scales. This mechanism effectively addresses the challenge of single structures being insufficient for fully extracting sequential features.

The forward process of DSAT-HD can be formulated as follows:

\begin{align}
\mathbf{X}^{\text{norm}} &= \text{InstanceNorm}(\mathbf{X}) \\
{\text{S}}, {\text{T}}& = \text{EMA}(\mathbf{X}^{\text{norm}}) \\
\mathbf{X}' &= \text{ChannelChange}(\mathbf{X}^{\text{norm}}) \\
\mathbf{X}_{\text{Freq}} &= \text{FFT}(\mathbf{X}') \\
{\text{S}}^{\text{Freq}}, {\text{T}}^{\text{Freq}} &= \text{DecompFreq}(\mathbf{X}_{\text{Freq}}) \\
\mathbf{G} &= \text{TopK}(\text{S}^{\text{Freq}}, \text{T}^{\text{Freq}}) \\
\mathbf{H} &= \text{SparseDispatcher}(\mathbf{X}', \mathbf{G}) \\
\mathbf{O} &= \text{MultiScaleExperts}(\mathbf{H}) \\
\hat{\mathbf{X}} &= \text{SparseCombiner}(\mathbf{O}, \mathbf{G})
\end{align}

\subsection{Hybrid Decomposition Module}
The hybrid decomposition module consists of three components: EMA decomposition, Fourier decomposition, and multi-scale moving average. EMA decomposition captures trends in the time domain, Fourier decomposition extracts global periodic features in the frequency domain, and the multi-scale moving average captures multi-scale features through convolutional kernels of different sizes.

Given an input sequence X, the hybrid decomposition process is as follows:

\begin{align}
\mathbf{X}_T^{\text{EMA}} &= \text{EMA}(\mathbf{X}) \\
\mathbf{X}_S^{\text{EMA}} &= \mathbf{X} - \mathbf{X}_T^{\text{EMA}} \\
\mathbf{X}_F &= \mathcal{FFT}(\mathbf{X})\\
\mathbf{X}_S^{\text{Freq}} &= \mathcal{FFT}^{-1}(\text{TopK}(\mathbf{X}_F)) \\
\mathbf{X}_T^{\text{Freq}} &=  \text{Trend}(\mathbf{X})\\
\mathbf{X}_T^{\text{EMA}} &=\text{MLPStream}(\mathbf{X_T^{EMA}}) \\
\mathbf{X}_S^{\text{EMA}} &= \text{CNNStream}(\mathbf{X_S^{EMA}}) \\
\end{align}

The multi-expert decomposition and fusion process based on the gating coefficients obtained from TopK is as follows (where COV denotes the coefficient of variation):

\begin{align}
    \text{Gates}, \text{L} &= \text{TopK}(\mathbf{X_F})\\
    \text{loss} &= \text{COV}(\sum \text{Gates}) + \text{COV}(\text{L})\\
    \text{Z} &=  \text{dispatcher}(X) \\
    \mathbf{X}_T^{Freq} &= \text{Fusion}_{i=1}^4(\text{Expert[i](Z[i])}) 
\end{align}

The final trend and seasonal components are fused, where [·;·] denotes the concatenation operation:

\begin{align}
\mathbf{G} &= \mathbf{W}_g [\mathbf{X}_T^{\text{EMA}};\mathbf{X}_S^{\text{EMA}}; \mathbf{X}_T^{\text{Freq}}] + \mathbf{b}_g\\
\end{align}

\subsection{Multi-scale Adaptive Path}

Fig.~\ref{fig:Transformer_layer} illustrates how the multi-scale adaptive path dynamically routes input features to four Transformer expert layers with different scales through a sparse dispatcher. Each expert layer is configured with a specific patch size to handle patterns at particular temporal scales.

Let the set of experts be denoted as $\{Expert_k\}_{k=1}^4$, with corresponding patch sizes $\{p_k\}_{k=1}^4$. The routing gating is computed as follows:

\begin{align}
    \mathbf{g},L &= \text{TopK}(\mathbf{X_F}) \\
    \mathbf{H}_k &= \text{Expert}_k(\mathbf{X}), \quad \text{for } k=1,\dots,4
\end{align}

The final output is integrated via the sparse combiner:

\begin{align}
    \mathbf{O} = \sum_{k=1}^{4} g_k \cdot \mathbf{H}_k
\end{align}

Each expert layer divides the sequence sequentially into  
patch $\frac{seq\_len}{patch}$ segments based on the patch size, iteratively performs intra-sequence fusion and self-attention mechanisms. For inter-sequence interactions, three shared weight matrices project all input sequences into $Q$, $K$, and $V$ to compute scaled dot-product attention. Finally, the features derived from intra-sequence and inter-sequence operations are fused.

\begin{align}
    H_{f}(i) &= \textbf{X}[i*patch:(i+1)patch] \\
    H_{L}(i)  &= \text{Fusion}_{i=1}^{\frac{seq\_len}{patch}}  (H_{f}(i), H_{f}(i-1)) \\
    H_{L}(i) &= SelfAttention(H_{L}(i))\\
    Q &=  Proj_Q(\textbf{X})\\
    K &=  Proj_K(\textbf{X})\\
    V &=  Proj_V(\textbf{X})\\
    H_{R} &=  Softmax(\frac{Q\times K^T}{\sqrt{k}})V\\
    H &= Fusion(H_{L}(\frac{seq\_len}{patch}), H_{R})
\end{align}

\subsection{Dual-Stream Residual Learning Framework}

Fig.~\ref{fig:model_overall} illustrates the dual-stream framework comprising two parallel branches: a CNN stream processes seasonal components, while an MLP stream handles trend components.

The CNN branch employs a hierarchical convolutional structure to capture multi-scale seasonal patterns:

\begin{align}
\mathbf{H}_s^0 &= \mathbf{X}_S \\
\mathbf{H}_s^l &= \text{ConvBlock}(\mathbf{H}_s^{l-1}), \quad l = 1,\dots,L
\end{align}

The MLP branch captures trend variations through fully connected layers:

\begin{align}
\mathbf{H}_t^0 &= \mathbf{X}_T \\
\mathbf{H}_t^l &= \text{MLPBlock}(\mathbf{H}_t^{l-1}), \quad l = 1,\dots,L
\end{align}

\subsection{Loss Function}
\begin{table*}[htbp]
    \centering
    \caption{Comparison of Forecasting Errors Between Baseline Models and the DSAT-HD Model Under a Uniform Lookback Window: All Datasets Employ a Window Length of $L=336$.}
    \resizebox{\textwidth}{!}{
    \begin{tabular}{c|c|cc|cc|cc|cc|cc|cc|cc|cc|cc|cc}
    \hline
    \multicolumn{2}{c}{\textbf{Method}}&  \multicolumn{2}{c}{\makecell{\textbf{DSAT-HD}\\(ours)}}&\multicolumn{2}{c}{\makecell{\textbf{xPatch}\\(AAAI2025)}} & \multicolumn{2}{c}{\makecell{\textbf{SDE-Mamba}\\(KDD2025)}} & \multicolumn{2}{c}{\makecell{\textbf{Pathformer}\\(ICLR2025)}} & \multicolumn{2}{c}{\makecell{\textbf{CASA}\\(ICLR2025)}} & \multicolumn{2}{c}{\makecell{\textbf{Amplifier}\\(AAAI2025)}} & \multicolumn{2}{c}{\makecell{\textbf{TimeMixer++}\\(ICLR2025)}} &   \multicolumn{2}{c}{\makecell{\textbf{SDE-SegRNN}\\(KDD2025)}
}&  \multicolumn{2}{c}{\makecell{\textbf{SimpleTM}\\(ICLR2025)}
}&\multicolumn{2}{c}{\makecell{\textbf{iTransformer}\\(ICLR2024)}
}\\
    \hline
    \multicolumn{2}{c}{\textbf{Metrix}}&\textbf{MSE}&\textbf{MAE}&\textbf{MSE}&\textbf{MAE}& \textbf{MSE}&  \textbf{MAE}&\textbf{MSE}&\textbf{MAE}& \textbf{MSE}&\textbf{MAE}& \textbf{MSE}& \textbf{MAE}& \textbf{MSE}&\textbf{MAE} & \textbf{MSE}&\textbf{MAE}
&  \textbf{MSE}&\textbf{MAE}
&\textbf{MSE}&\textbf{MAE}
\\ 
    \hline
  \multirow{4}{*}{\rotatebox[origin=c]{90}{\textbf{ETTh1}}}&96& \textbf{0.357}& \textbf{0.381}& 0.376& \underline{0.386}& 0.376&  0.400&0.382&0.400 & 0.376&0.397& 0.371& 0.392& \underline{0.361}&0.403 &   0.376&0.400&  0.366&0.392&0.386&0.405
\\
  &192& \textbf{0.377}& \textbf{0.396}& 0.417& \underline{0.407}& 0.432&  0.429&0.440& 0.427& 0.427& 0.424& 0.426& 0.422& \underline{0.416}&0.441 &   0.432&0.429&  0.422&0.421&0.424&0.440
\\
  &336& \textbf{0.420}& \textbf{0.422}& 0.449& \underline{0.425}& 0.477&  0.437&0.454& 0.432& 0.469& 0.445& 0.448& 0.434& \underline{0.430}&0.434 &   0.477&0.437&  0.440&0.438&0.449&0.460
\\
  &720& \textbf{0.443}& \underline{0.458}& 0.470& 0.459& 0.488 &  0.471&0.479& 0.461& 0.479& 0.468& 0.476& 0.464& \underline{0.467}&\textbf{0.451} &   0.488&0.471&  0.463&0.462&0.495&0.487
\\
  \hline
 \multirow{4}{*}{\rotatebox[origin=c]{90}{\textbf{ETTh2}}}& 96& \textbf{0.223}& \textbf{0.297}& \underline{0.233}&
\underline{0.300}& 0.288&  0.340&0.279& 0.331& 0.290& 0.339& 0.279& 0.337& 0.276 &0.328 &   0.288&0.340&  0.281&0.338&0.297&0.348
\\
 & 192& \textbf{0.272}& \textbf{0.331}& \underline{0.291}&
\underline{0.338}& 0.373&  0.390&0.349& 0.380& 0.366& 0.388& 0.359& 0.389& 0.342 &0.379 &   0.373&0.390&  0.355&0.387&0.372&0.403
\\
 & 336& \textbf{0.311}& \textbf{0.363}& \underline{0.344}&
\underline{0.377}& 0.380 &  0.406&0.348& 0.382& 0.416& 0.425& 0.377& 0.406& 0.346 &0.398 &   0.380&0.406&  0.365&0.401&0.388&0.417
\\
 & 720& \textbf{0.395}& \textbf{0.426}& 0.407&0.427 & 0.412 &  0.432&\underline{0.398}& \underline{0.424}& 0.420& 0.439& 0.420& 0.432& 0.392 &0.415 &   0.412 &0.432&  0.413 &0.436&0.424&0.444
\\
 \hline
 \multirow{4}{*}{\rotatebox[origin=c]{90}{\textbf{ETTm1}}}& 96& \textbf{0.287}& \textbf{0.332}& 0.311&
0.346 & 0.315&  0.357&0.316& 0.346& 0.313& 0.358& 0.316& 0.355& 0.310 &\underline{0.334} &   0.315&0.357&  0.321&0.361&\underline{0.300}&0.353
\\
 & 192& \textbf{0.332}& \textbf{0.358}& 0.348&
0.368& 0.360 &  0.383&0.366& 0.370& 0.375& 0.382& 0.361& 0.381& 0.348&\underline{0.362} &   0.360 &0.383&  0.360&0.380&\underline{0.341}&0.380
\\
 & 336& \textbf{0.357}& \textbf{0.379}& 0.388&
0.391 & 0.389 &  0.405&0.386& 0.394& 0.397& 0.398& 0.393& 0.402& 0.376&\underline{0.391} &   0.389&0.405&  0.390&0.404&\underline{0.374}&0.396
\\
 & 720& \textbf{0.422}& \textbf{0.412}& 0.461&0.430 & 0.448&  0.440&0.460& 0.432& 0.459& 0.437& 0.455& 0.440& 0.440 &\underline{0.423} &   0.448&0.440&  0.454&0.438&\underline{0.429}&0.430
\\
 \hline
 \multirow{4}{*}{\rotatebox[origin=c]{90}{\textbf{ETTm2}}}& 96& \textbf{0.149}& \textbf{0.239}& \underline{0.164}&
0.248& 0.172&  0.259&0.170& 0.248& 0.173& 0.252& 0.176& 0.258& 0.170 &\underline{0.245} &   0.172&0.259&  0.173&0.257&0.175&0.266
\\
 & 192& \textbf{0.219}& \textbf{0.285}& 0.230&
\underline{0.291}& 0.238&  0.301&0.238& 0.295& 0.240& 0.298& 0.239& 0.300& \underline{0.229}&0.291 &   0.238&0.301&  0.238&0.299&0.242&0.312
\\
 & 336& \textbf{0.266}& \textbf{0.318}& 0.292&
\underline{0.331}& 0.300&  0.340&0.293& 0.331& 0.296& 0.334& 0.297& 0.338& 0.303&0.343 &   0.300&0.340&  0.296&0.338&\underline{0.282}&0.337
\\
 & 720& \textbf{0.343}& \textbf{0.367}& 0.381&\underline{0.383}& 0.394&  0.394&0.390& 0.389& 0.395& 0.392& 0.393& 0.396& \underline{0.373}&0.399 &   0.394&0.394&  0.393&0.395&0.375&0.394
\\
 \hline
 \multirow{4}{*}{\rotatebox[origin=c]{90}{\textbf{Weather}}}& 96& \textbf{0.147}& \textbf{0.189}& 0.168&
0.203 & 0.165&  0.214&0.156& \underline{0.192}& 0.155& 0.197& 0.156& 0.204& \underline{0.155}&0.205 &   0.165& 0.214&  0.162&0.207&0.157&0.207
\\
 & 192& \textbf{0.189}& \textbf{0.227}& 0.214&
0.245 & 0.214&  0.255&0.206& \underline{0.240}& 0.206& 0.245& 0.209& 0.249& 0.201&0.245 &   0.214&0.255&  0.208& 0.248&\underline{0.200}&0.248
\\
 & 336& \underline{0.239}& \textbf{0.267}& \textbf{0.236}&
0.273& 0.271&  0.297&0.254& 0.282& 0.265 & 0.286& 0.264& 0.290& 0.239&\underline{0.267} &   0.271& 0.297& 0.263&0.290&0.252&0.287
\\
 & 720& \underline{0.316}& \underline{0.324}& \textbf{0.309}&\textbf{0.321}& 0.346 &  0.347&0.340& 0.336& 0.347 & 0.340& 0.343& 0.342& 0.316&0.335 &  0.346 & 0.347& 0.340& 0.341&0.320&0.336
\\
 \hline
 \multirow{4}{*}{\rotatebox[origin=c]{90}{\textbf{Traffic}}}& 96& \underline{0.375}& \underline{0.253}& 0.481&
0.280 & 0.388&  0.261&0.479& 0.283& 0.386& \textbf{0.243}& 0.455& 0.298& 0.392&0.253 & 0.388& 0.261&  0.410&0.274&\textbf{0.363}&0.265
\\
 & 192& \underline{0.398}& 0.264& 0.484&
0.275 & 0.411 &  0.271&0.484& 0.292& 0.410& \textbf{0.255}& 0.470& 0.316& 0.402&\underline{0.258} &   0.411& 0.271&  0.430&0.280&\textbf{0.384}&0.273
\\
 & 336& \underline{0.410}& \underline{0.270}& 0.504&
0.279 & 0.428 &  0.278&0.503& 0.299& 0.426& \textbf{0.263}& 0.479& 0.316& 0.428&0.273 & 0.428 &0.278&  0.449&0.290&\textbf{0.396}&0.277
\\
 & 720& \textbf{0.440}& \textbf{0.282}& 0.540&0.293& 0.461&  0.297& 0.537& 0.322& 0.461&  \underline{0.282}& 0.523& 0.328& \underline{0.441}& 0.282 &  0.461 & 0.297&  0.486& 0.309& 0.445& 0.308
\\
 \hline
 \multirow{4}{*}{\rotatebox[origin=c]{90}{\textbf{Electricity}}}& 96& \textbf{0.131}& \underline{0.225}& 0.159&
0.244 & 0.146&  0.244&0.145& 0.236& 0.138& 0.232& \underline{0.132}& \underline{0.227}& 0.135 &\textbf{0.222} &   0.146&0.244&  0.141&0.235&0.134&0.230
\\
 & 192& \underline{0.153}& 0.252& 0.160&
0.248& 0.162&  0.258&0.167& 0.256& 0.152& \underline{0.247}& 0.159& 0.252& \textbf{0.147}&\textbf{0.235} &  0.162 &0.258& 0.151 &0.247&0.154&0.250
\\
 & 336& \textbf{0.162}& \underline{0.256}& 0.182&
0.267 & 0.177&  0.274&0.186& 0.275& 0.208& 0.306& 0.214& 0.319& \underline{0.164}&\textbf{0.245} & 0.177  &0.274&  0.173&0.267&0.169&\underline{0.265}
\\
 & 720& \textbf{0.187}& \textbf{0.282}& 0.216&0.298 & 0.202&  0.297&0.231& 0.309& 0.253& 0.341& 0.257& 0.342& 0.212&0.310 &  0.202 &0.297&  0.201&0.293&\underline{0.194}&\underline{0.288}
\\
 \hline
 \multirow{4}{*}{\rotatebox[origin=c]{90}{\textbf{Exchange}}}& 96& \textbf{0.081}& \textbf{0.197}& \underline{0.082}&
\underline{0.199}& 0.083 &  0.202&0.088& 0.208& 0.113& 0.236& 0.083& 0.202& 0.085&0.214 &   0.083 &0.202&  0.091&0.213&0.086&0.205
\\
 & 192& \textbf{0.175}& \textbf{0.297}& 0.177&
\underline{0.298}& \underline{0.176}&  0.298&0.183& 0.304& 0.216& 0.333& 0.176& 0.297& \underline{0.175}&0.313 &   0.176&0.298&  0.194&0.316&0.177&0.299
\\
 & 336& 0.366& 0.435& 0.349&
0.425 & 0.327&  0.413&0.354& 0.429& 0.649& 0.651& 0.328& 0.414& \textbf{0.316}&\underline{0.420} &   0.327&0.413&  0.373&0.447&\underline{0.331}&\textbf{0.417}
\\
 & 720& 1.070& 0.770& 0.891&0.711 & \textbf{0.839}&  0.689&0.909& 0.716& 1.082& 0.78& 0.858& 0.696& 0.851&\textbf{0.689} &   \underline{0.839} &0.689&  1.111&1.216&0.846&\underline{0.693}
\\
  \hline
 \multirow{4}{*}{\rotatebox[origin=c]{90}{\textbf{Solar}}}& 96
& \textbf{0.162}& \textbf{0.196}& 0.194&
 \underline{0.209}& 0.186&  0.217&0.218& 0.235& 0.187& 0.218& 0.174& 0.237& 0.171&0.231 & 0.186& 0.217&  \underline{0.163}&0.232&0.190&0.244
\\
 & 192
& \textbf{0.183}& \textbf{0.208}& 0.234&
 0.235& 0.230&  0.251&0.196& \underline{0.220}& 0.223& 0.243& 0.197& 0.259& 0.218&0.263 & 0.230& 0.251&  \underline{0.183}&0.247&0.193&0.257
\\
 & 336
& \underline{0.200}& \textbf{0.216}& 0.256&
 0.253& 0.253&  0.270&0.202& 0.220& 0.237& 0.257& 0.206& 0.263& 0.212&0.269 &   0.253& 0.270&  \textbf{0.193}&0.257&0.203&0.266
\\
 & 720& \underline{0.203}& \textbf{0.217}& 0.262& 0.257& 0.247&  0.274&0.208& \underline{0.237}& 0.239& 0.258& 0.222& 0.268& 0.212&0.270 & 0.247&0.274&  \textbf{0.199}&0.252&0.223&0.281\\
    \hline
    \multicolumn{2}{c|}{\textbf{$1^{ST}\text{ Count}$}} & 26 & 26 & 2 & 1 & 1 & 0 & 0 & 0 & 0 & 3 & 0 & 0 & 2 & 5 & 0 & 0 & 2 & 0 &  3 & 1
    \\
    \hline
    \end{tabular}}
    \label{tab:main_result}
\end{table*}

\begin{table*}[h]
    \centering
    \caption{Input length \textit{H} = 48, forecast horizon $F\in\{96,192,336,720\}$, with best results highlighted in bold.}
    \begin{tabular}{c|c|cc|cc|cc|cc}
    \hline
    \multicolumn{2}{c}{\textbf{Method}}&  \multicolumn{2}{c}{\textbf{DSAT-HD}}&\multicolumn{2}{c}{\textbf{PatchTST}} & \multicolumn{2}{c}{\textbf{Autoformer}} & \multicolumn{2}{c}{\textbf{FEDformer}}   \\
    \hline
    \multicolumn{2}{c}{\textbf{Metrix}}&\textbf{MSE}&\textbf{MAE}&\textbf{MSE}&\textbf{MAE}& \textbf{MSE}&  \textbf{MAE}&\textbf{MSE}&\textbf{MAE}\\ 
    \hline
  \multirow{4}{*}{\textbf{ETTh1}}&96& \underline{0.390}& \textbf{0.393}& 0.410& \textbf{0.417}& 0.406&  0.432&\textbf{0.382}&0.419\\
  &192& \textbf{0.433}& \textbf{0.413}& 0.469& \underline{0.448}& \underline{0.451}&  0.452&0.451& 0.456\\
  &336& \underline{0.469}& \textbf{0.434}& 0.516& \underline{0.469}& \textbf{0.461}&  0.464&0.499& 0.487\\
  &720& \textbf{0.480}& \textbf{0.473}& 0.509& \underline{0.487}& \underline{0.498}&  0.500&0.510& 0.504\\
  \hline
 \multirow{4}{*}{\textbf{ETTh2}}& 96& \textbf{0.245}& \textbf{0.304}& \underline{0.307}&
\underline{0.348}& 0.344&  0.383&0.330& 0.373\\
 & 192& \textbf{0.305}& \textbf{0.345}& \underline{0.397}&
\underline{0.399}& 0.425&  0.426&0.440& 0.436\\
 & 336& \textbf{0.358}& \textbf{0.383}& \underline{0.399}&
\underline{0.420}& 0.445&  0.452&0.543& 0.504\\
 & 720& \textbf{0.423}& \textbf{0.433}& \underline{0.434}&\underline{0.441}& 0.483&  0.481&0.471& 0.483\\
 \hline
 \multirow{4}{*}{\textbf{ETTm1}}& 96& \textbf{0.415}& \textbf{0.393}& \underline{0.424}&
\underline{0.403}& 0.745&  0.556&0.428& 0.432\\
 & 192& \textbf{0.452}& \textbf{0.414}& \underline{0.468}&
\underline{0.429}& 0.715&  0.556&0.476& 0.460\\
 & 336& \textbf{0.492}& \textbf{0.440}& \underline{0.501}&
\underline{0.453}& 0.816&  0.590&0.526& 0.494\\
 & 720& \textbf{0.552}& \textbf{0.437}& \underline{0.553}&\underline{0.484}& 0.746&  0.572&0.630& 0.528\\
 \hline
 \multirow{4}{*}{\textbf{ETTm2}}& 96& \textbf{0.172}& \textbf{0.254}& 0.189&
\underline{0.272}& 0.211&  0.299&\underline{0.185}& 0.274\\
 & 192& \textbf{0.237}& \textbf{0.296}& 0.260&
0.371& 0.277&  0.388&\underline{0.256}& \underline{0.318}\\
 & 336& \textbf{0.305}& \textbf{0.340}& \underline{0.328}&
\underline{0.359}& 0.347&  0.380&0.329& 0.365\\
 & 720& \textbf{0.396}& \textbf{0.396}& \underline{0.429}&\underline{0.415}& 0.441&  0.432&0.447& 0.432\\
 \hline
 \multirow{4}{*}{\textbf{Weather}}& 96& \textbf{0.195}& \textbf{0.225}& \underline{0.212}&
\underline{0.243}& 0.291&  0.357&0.241& 0.309\\
 & 192& \textbf{0.229}& \textbf{0.257}& \underline{0.254}&
\underline{0.277}& 0.349&  0.391&0.308& 0.356\\
 & 336& \textbf{0.258}& \textbf{0.289}& \underline{0.310}&
\underline{0.316}& 0.409&  0.424&0.385& 0.406\\
 & 720& \textbf{0.339}& \textbf{0.342}& \underline{0.385}&\underline{0.365}& 0.437&  0.431&0.438& 0.432\\
 \hline
 \multirow{4}{*}{\textbf{Electricity}}& 96& \underline{0.218}& \textbf{0.282}& 0.225&
\underline{0.293}& \textbf{0.211}&  0.322&0.240& 0.349\\
 & 192& \textbf{0.211}& \textbf{0.282}& 0.229&
\underline{0.299}& \underline{0.224}&  0.331&0.248& 0.357\\
 & 336& \textbf{0.214}& \textbf{0.293}& \underline{0.239}&
\underline{0.316}& 0.259&  0.362&0.265& 0.370\\
 & 720& \textbf{0.217}& \textbf{0.332}& 0.282&0.349& 0.313&  0.407&0.326& 0.405
 \\
 \hline
    \end{tabular}
    \label{tab:result_48}
\end{table*}
The total loss function is composed of the prediction loss and the balance loss:

\begin{align}
\mathcal{L} = \mathcal{L}_{\text{pred}} + \mathcal{L}_{\text{balance}}
\end{align}

The balance loss function ensures balanced collaboration among experts:

\begin{align}
g, \text{load} &= \text{Topk}(\mathbf{X})\\
\text{g}_{\text{sum}} &= \sum g \\
\mathcal{L}_{\text{balance}} &= \text{COV}(\text{g}_{\text{sum}}) + \text{COV}(\text{load})
\end{align}

Where $L_{pred}=\|\hat{\text{X}}-\text{Y}\|_2^2$ denotes the Mean Squared Error (MSE) loss.

\section{Experiments and Results}

\begin{table*}[htbp]
    \centering
    \caption{Ablation Study: w/o Hybrid Decomposition, w/o Multi-scale Path (MoE), and w/o Dual-Stream Framework denote the removal of the EMA and Fourier decomposition modules, the multi-scale routing mechanism, and the dual-stream (CNN+MLP) framework, respectively. Results are averaged across all forecasting horizons.}
    \begin{tabular}{c|cc|cc|cc|cc}
    \hline
    Method&  \multicolumn{2}{c}{ETTh2}&\multicolumn{2}{c}{ETTm2} & \multicolumn{2}{c}{Weather} & \multicolumn{2}{c}{Traffic}   \\
    \hline
    Metrix&MSE&MAE&MSE&MAE & MSE&  MAE&MSE&MAE \\ 
    \hline
    MSAT-HD& \textbf{0.300}& \textbf{0.354}& \textbf{0.244}& \textbf{0.302}& \textbf{0.222}&  \textbf{0.251}&\textbf{0.406}&\textbf{0.268}\\
    \hline
  w/o Hybrid Decomposition& 0.306& 0.358& 0.247& 0.314& 0.227&  0.255&0.410& 0.275\\
  \hline
  w/o Multi-scale Path (MoE)& 0.318& 0.373& 0.266& 0.313& 0.231&  0.260&0.502& 0.281\\
  \hline
  w/o Dual-Stream Framework& 0.317& 0.369& 0.254& 0.308& 0.248&  0.277&0.701& 0.405\\
    \hline
    \end{tabular}
    \label{tab:ablation study}
\end{table*}

\subsection{Datasets}

\begin{table*}[h]
    \centering
    \caption{Input length \textit{H} = 192, forecast horizon $F\in\{96,192,336,720\}$, with best results highlighted in bold.}
    \begin{tabular}{c|c|cc|cc|cc|cc}
    \hline
    \multicolumn{2}{c}{\textbf{Method}}&  \multicolumn{2}{c}{\textbf{DSAT-HD}}&\multicolumn{2}{c}{\textbf{PatchTST}} & \multicolumn{2}{c}{\textbf{Autoformer}} & \multicolumn{2}{c}{\textbf{FEDformer}}   \\
    \hline
    \multicolumn{2}{c}{\textbf{Metrix}}&\textbf{MSE}&\textbf{MAE}&\textbf{MSE}&\textbf{MAE}& \textbf{MSE}&  \textbf{MAE}&\textbf{MSE}&\textbf{MAE}\\ 
    \hline
  \multirow{4}{*}{\textbf{ETTh1}}&96& \textbf{0.372}& \textbf{0.384}& \underline{0.384}& \underline{0.403}& 0.430&  0.441&0.388&0.423\\
  &192& \textbf{0.404}& \textbf{0.399}& \underline{0.428}& \underline{0.425}& 0.487&  0.467&0.433& 0.456\\
  &336& \textbf{0.430}& \textbf{0.418}& 0.452& \underline{0.436}& 0.478&  0.474&\underline{0.445}& 0.462\\
  &720& 0.478& \textbf{0.459}& \textbf{0.453}& \underline{0.459}& 0.518&  0.519&\underline{0.476}& 0.490\\
  \hline
 \multirow{4}{*}{\textbf{ETTh2}}& 96& \textbf{0.234}& \textbf{0.301}& \underline{0.285}&
\underline{0.340}& 0.362&  0.401&0.397& 0.424\\
 & 192& \textbf{0.284}& \textbf{0.334}& \underline{0.356}&
\underline{0.387}& 0.430&  0.447&0.439& 0.458\\
 & 336& \textbf{0.330}& \textbf{0.369}& \underline{0.351}&
\underline{0.396}& 0.408&  0.447&0.471& 0.481\\
 & 720& \underline{0.404}& \textbf{0.427}& \textbf{0.395}&\underline{0.427}& 0.440&  0.469&0.479& 0.490\\
 \hline
 \multirow{4}{*}{\textbf{ETTm1}}& 96& \textbf{0.281}& \textbf{0.332}& \underline{0.295}&
\underline{0.345}& 0.510&  0.428&0.381& 0.424\\
 & 192& \textbf{0.317}& \textbf{0.355}& \underline{0.330}&
\underline{0.365}& 0.619&  0.545&0.412& 0.441\\
 & 336& \textbf{0.354}& \textbf{0.379}& \underline{0.364}&
\underline{0.388}& 0.561&  0.500&0.435& 0.455\\
 & 720& \underline{0.428}& \textbf{0.420}& \textbf{0.423}&\underline{0.424}& 0.580&  0.512&0.473& 0.474\\
 \hline
 \multirow{4}{*}{\textbf{ETTm2}}& 96& \textbf{0.156}& \textbf{0.241}& \underline{0.169}&
\underline{0.254}& 0.244&  0.321&0.223& 0.305\\
 & 192& \textbf{0.216}& \textbf{0.280}& \underline{0.230}&
\underline{0.294}& 0.302&  0.362&0.281& 0.339\\
 & 336& \textbf{0.269}& \textbf{0.317}& \underline{0.281}&
\underline{0.329}& 0.346&  0.390&0.321& 0.364\\
 & 720& \textbf{0.350}& \textbf{0.370}& \underline{0.373}&\underline{0.384}& 0.423&  0.428&0.417& 0.420\\
 \hline
 \multirow{4}{*}{\textbf{Weather}}& 96& \textbf{0.148}& \textbf{0.185}& \underline{0.160}&
\underline{0.205}& 0.298&  0.363&0.239& 0.316\\
 & 192& \textbf{0.197}& \textbf{0.232}& \underline{0.204}&
\underline{0.245}& 0.322&  0.379&0.274& 0.326\\
 & 336& \textbf{0.224}& \textbf{0.262}& \underline{0.258}&
\underline{0.285}& 0.378&  0.409&0.334& 0.369\\
 & 720& \textbf{0.299}& \textbf{0.315}& \underline{0.329}&\underline{0.337}& 0.435&  0.431&0.401& 0.412\\
 \hline
 \multirow{4}{*}{\textbf{Electricity}}& 96& \textbf{0.138}& \textbf{0.231}& \underline{0.146}&
\underline{0.240}& 0.198&  0.313&0.231& 0.343\\
 & 192& \textbf{0.147}& \textbf{0.242}& \underline{0.152}&
\underline{0.252}& 0.218&  0.335&0.258& 0.361\\
 & 336& \textbf{0.169}& \textbf{0.263}& \underline{0.178}&
\underline{0.271}& 0.252&  0.352&0.273& 0.372\\
 & 720& \textbf{0.216}& \textbf{0.298}& \underline{0.223}&\underline{0.308}& 0.275&  0.371&0.308& 0.402\\ \hline
    \end{tabular}
    \label{tab:result_192}
\end{table*}

\begin{table*}[htbp]
    \caption{Dataset Statistics}
    \centering
    \begin{tabular}{cccccccc}
        \hline
         \textbf{Datasets}&  \textbf{ETTh1\&ETTh2}&  \textbf{ETTm1\&ETTm2}&  \textbf{Weather}&  \textbf{Electricity}&  \textbf{Traffic}&  \textbf{Exchange}& \textbf{Solar}\\
        \hline 
         \textbf{Variables}&  7&  7&  21&  321&  862&  8& 137\\
         \textbf{Timestamps}&  17420&  69680&  52696&  26304&  17544&  7207& 52603\\
         \textbf{Split Ratio}& 6:2:2& 6:2:2& 7:1:2& 7:1:2& 7:1:2& 7:1:2&7:1:2\\
        \hline
    \end{tabular}
    \label{tab:数据集统计}
\end{table*}

We evaluate the performance of DSAT-HHD on nine public datasets, including the benchmark datasets ETTh1, ETTh2, ETTm1, ETTm2, Electricity, Traffic, Weather, Solar, and Exchange Rate. All experiments are implemented using the PyTorch framework. The model is trained using the AdamW optimizer with an initial learning rate of $5\times10^{-5}$.

\begin{enumerate}
    \item ETT\footnote{https://github.com/zhouhaoyi/ETDataset}: It contains four subsets, where ETTh1 and ETTh2 are recorded at an hourly frequency, while ETTm1 and ETTm2 are recorded at a 15-minute interval. The data are collected from two different electricity transformers.
    \item Weather\footnote{https://www.bgc-jena.mpg.de/wetter/}: This dataset includes 21 meteorological indicators (e.g., humidity and temperature) in Germany, recorded every 10 minutes by a weather station at the Max Planck Institute for Biogeochemistry during the year 2020.
    \item Traffic\footnote{https://pems.dot.ca.gov/}: It records the hourly road occupancy rates measured by 862 sensors on highways in the San Francisco Bay Area from January 2015 to December 2016.
    \item Electricity\footnote{https://archive.ics.uci.edu/ml/datasets/ElectricityLoadDiagrams20112014}: The electricity dataset contains hourly electricity consumption records of 321 clients from July 2016 to July 2019.
    \item Exchange-rate\footnote{https://github.com/laiguokun/multivariate-time-series-data}: This dataset collects the daily exchange rate panel data of eight countries from 1990 to 2016.
    \item Solar-energy\footnote{https://www.nrel.gov/grid/solar-power-data.html}: It contains solar power generation records from 137 photovoltaic power plants in the year 2006.
\end{enumerate}

During data preparation, we follow the established methodologies from previous studies \cite{wu2022timesnet,wu2021autoformer}. Detailed statistics are summarized in Table \ref{tab:数据集统计}.

\subsection{Impact of Varying Input Lengths on the Model}
In time series forecasting, the input length directly influences the amount of historical information available to the model. To evaluate the effectiveness of DSAT-HD, we tested its performance under different input lengths and visualized the forecasting results at input lengths of 48 and 192 (Fig. \ref{fig:48和192长度比较}). On the ETTh1, ETTh2, Weather, and Electricity datasets, the DSAT-HD model consistently outperformed the baselines on some datasets. As shown in Table \ref{tab:result_48} and Table \ref{tab:result_192}, when the forecast horizon H=48, the model achieved optimal performance in 45 out of 48 tests; similarly, when H=192, it also attained optimal results in 45 out of 48 tests. These findings indicate that DSAT-HD exhibits exceptional generalization capabilities across varying input lengths, with its performance continuously improving as the input length increases, thereby demonstrating its capability to effectively leverage long-sequence information.

\subsection{Baselines}

In the field of time series forecasting, numerous models have emerged in recent years. We select baseline models demonstrating exceptional forecasting performance between 2024 and 2025, including xPatch (AAAI 2025), SDE-Mamba (KDD 2025), Pathformer (ICLR 2025), CASA (ICLR2025), Amplifier (AAAI2025), TimeMixer++ (ICLR2025), SDE-SegRNN (KDD2025), SimpleTM (ICLR2025), among others, as well as state-of-the-art (SOTA) models from 2024 such as iTransformer. The code repositories for these models are as follows:

\begin{itemize}
    \item xPatch: https://github.com/stitsyuk/xPatch
    \item SDE-Mamba: https://github.com/YukinoAsuna/SAMBA
    \item Pathformer: https://github.com/decisionintelligence/pathformer
    \item CASA: https://github.com/lmh9507/CASA
    \item Amplifier: https://github.com/aikunyi/Amplifier
    \item TimeMixer++: https://github.com/kwuking/TimeMixer
    \item SDE-SegRNN: https://github.com/YukinoAsuna/SAMBA
    \item SimpleTM: https://github.com/vsingh-group/SimpleTM
    \item iTransformer: https://github.com/thuml/iTransformer
\end{itemize}

\subsection{Comparative Experiments}
Table \ref{tab:main_result} presents the comparative results between DSAT-HD and mainstream time series forecasting methods, including baseline approaches such as xPatch, SDE-Mamba, Pathformer, CASA, Amplifier, TimeMixer++, SDE-SegRNN, SimpleTM and iTransformer. The evaluation metrics include Mean Absolute Error (MAE) and Mean Squared Error (MSE).

\subsection{Ablation Studies}
To determine the impact of different modules within DSAT-HD, we conduct ablative experiments focusing on the following components:
\begin{enumerate}
    \item $w/o\ Hybrid\ Decomposition$: Removes both the EMA and Fourier decomposition modules.
    \item $w/o\ Multi\text{-}scale\ Path(MoE)$: Removes the multi-scale routing mechanism.
    \item $w/o\ Dual\text{-}Stream\ Framework$: Removes the dual-stream (CNN+MLP) module.
\end{enumerate}

Table \ref{tab:ablation study} shows the individual impact of each module. We derive the following observations: 

\begin{enumerate}
    \item When the EMA and Fourier decomposition modules are removed, the performance degradation is relatively minor on datasets with less distribution shift (e.g., Weather and Traffic). However, a significant performance drop is observed on datasets with substantial temporal variations (e.g., ETTm2), demonstrating the effectiveness of the EMA and Fourier decomposition modules. 
    
    \item Removing the multi-scale routing mechanism leads to noticeable performance degradation across all datasets, highlighting the necessity of the multi-scale routing module. 
    
    \item Similarly, removing the dual-stream module results in significant performance declines on all datasets, underscoring the essential role of the dual-stream framework.
\end{enumerate}

\subsection{Visualization}
\begin{figure*}[htbp]
\centering
\includegraphics[width=0.75\linewidth]{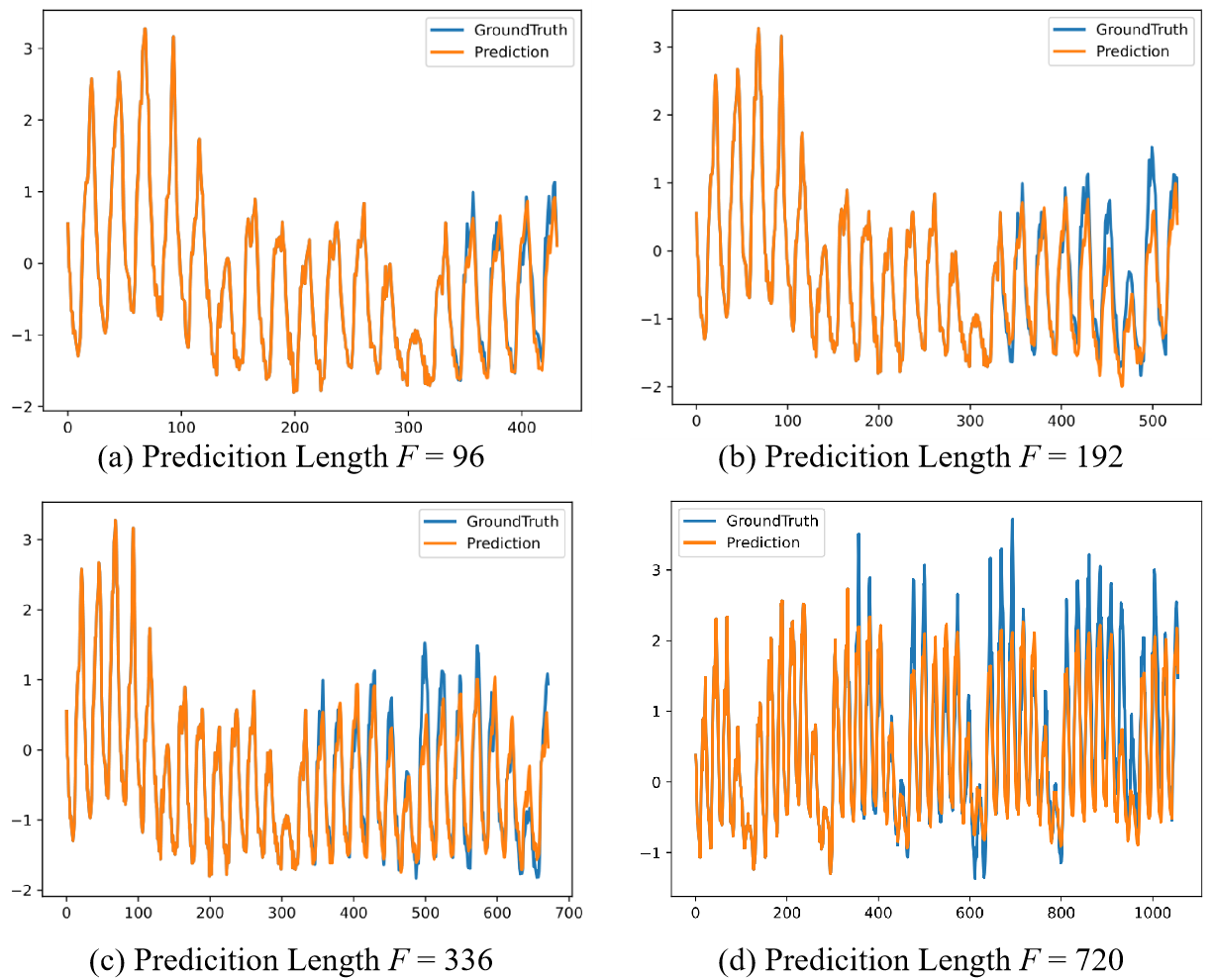}
\caption{Forecasting Results of the DSAT-HD Model on the Electricity Dataset with Input Length \textit{H} = 96}
\label{fig:ele_predict}
\end{figure*}

We present a visual comparison of the forecasting results generated by the DSAT-HD model on the Electricity dataset. As illustrated in Fig.~\ref{fig:ele_predict}, the predicted curves exhibit a high degree of alignment with the ground truth values across different prediction horizons (F=96,192,336,720). This demonstrates the exceptional forecasting capability of DSAT-HD. Furthermore, the model effectively captures complex trend and seasonal patterns present across multiple periods within diverse samples, providing evidence for its adaptive modeling capacity of multi-scale features.

\section{Conclusion}

This paper proposes DSAT-HD, a hybrid decomposition dual-stream adaptive Transformer framework for multivariate time series forecasting. By integrating a hybrid decomposition mechanism, a multi-scale adaptive path, and dual-stream residual learning, DSAT-HD effectively captures complex temporal dependencies and multi-scale characteristics in time series. Experimental results on multiple benchmark datasets demonstrate that DSAT-HD achieves overall superior performance in both forecasting accuracy and generalization capability. Future work will explore more efficient gating mechanisms and extend the framework to larger-scale datasets.

\section*{Acknowledgment}

This work is supported by the National Natural Science Foundation of China under Grant No. 82374621. The authors would also like to acknowledge the Traditional Chinese Medicine Data Center, China Academy of Chinese Medical Sciences, Beijing, China (contact: zhangxp@ndctcm.cn) for providing the data resources essential for this research.


\vspace{12pt}

\end{document}